\def\year{2021}\relax
\documentclass[letterpaper]{article}
\usepackage{arxiv} 
\usepackage{times} 
\usepackage{helvet} 
\usepackage{courier} 
\usepackage[hyphens]{url} 
\usepackage{graphicx} 
\usepackage{graphicx}  
\usepackage{natbib}
\usepackage{caption}
\usepackage{times}
\usepackage{latexsym}
\usepackage{graphicx}  
\usepackage{subfigure}
\usepackage{url}
\usepackage{amsfonts} 
\usepackage{amsmath}
\usepackage{booktabs}
\usepackage{multirow}
\usepackage{arydshln}
\usepackage{booktabs}
\usepackage{pifont}
\newcommand{\xmark}{\ding{55}}%

\title{Filling the Gap of Utterance-aware and Speaker-aware Representation for Multi-turn Dialogue}
\author{
    Longxiang Liu\textsuperscript{\rm 1,2,3,\thanks{Equal contribution. $\dagger$Corresponding author. This paper was partially supported by National Key Research and Development Program of China (No. 2017YFB0304100), Key Projects of National Natural Science Foundation of China (U1836222 and 61733011), Huawei-SJTU long term AI project, Cutting-edge Machine reading comprehension and language model.}},
	Zhuosheng Zhang\textsuperscript{\rm 1,2,3,*},
	Hai Zhao\textsuperscript{\rm 1,2,3,$\dagger$},
    Xi Zhou\textsuperscript{\rm 4},
    Xiang Zhou\textsuperscript{\rm 4}
	\\
}
\affiliations{
    \textsuperscript{\rm 1}Department of Computer Science and Engineering, Shanghai Jiao Tong University\\
	\textsuperscript{\rm 2}Key Laboratory of Shanghai Education Commission for Intelligent Interaction\\
	and Cognitive Engineering, Shanghai Jiao Tong University, Shanghai, China\\
	\textsuperscript{\rm 3}MoE Key Lab of Artificial Intelligence, AI Institute, Shanghai Jiao Tong University, Shanghai, China\\
	\textsuperscript{\rm 4}CloudWalk Technology, Shanghai, China\\
	{\tt \{com\_prehensive, zhangzs\}@sjtu.edu.cn, zhaohai@cs.sjtu.edu.cn,  \{zhouxi,zhouxiang\}@cloudwalk.cn}
}
\begin{document}
\maketitle
\begin{abstract}
A multi-turn dialogue is composed of multiple utterances from two or more different speaker roles. Thus utterance- and speaker-aware clues are supposed to be well captured in models. However, in the existing retrieval-based multi-turn dialogue modeling, the pre-trained language models (PrLMs) as encoder represent the dialogues coarsely by taking the pairwise dialogue history and candidate response as a whole, the hierarchical information on either utterance interrelation or speaker roles coupled in such representations is not well addressed. In this work, we propose a novel model to fill such a gap by modeling the effective utterance-aware and speaker-aware representations entailed in a dialogue history. In detail, we decouple the contextualized word representations by masking mechanisms in Transformer-based PrLM, making each word only focus on the words in current utterance, other utterances, two speaker roles (i.e., utterances of sender and utterances of receiver), respectively. Experimental results show that our method boosts the strong ELECTRA baseline substantially in four public benchmark datasets, and achieves various new state-of-the-art performance over previous methods. A series of ablation studies are conducted to demonstrate the effectiveness of our method.
\end{abstract}

\section{Introduction}
Building an intelligent chatbot that can understand human conversations and give logically correct, fluent responses is one of the eternal goals of artificial intelligence, especially in the field of natural language processing (NLP). The methods of building a chatbot which is capable of performing multi-turn dialogue can be categorized into two lines: generation-based methods and retrieval-based methods. Generation-based methods \cite{shang2015neural,serban2015building,su2019improving,zhang2020neural,li2019data,zhao-etal-2020-knowledge-grounded,shen-feng-2020-cdl} directly generate a response using an encoder-decoder framework, which tends to be short and less varied. Retrieval-based methods \cite{wu2016sequential,tao2019multi,tao2019one,zhang2018modeling,zhou2018multi,yuan2019multi,li58deep,hua2020learning} retrieve a list of response candidates, then use a matching model to rerank the candidates and select the best one as a reply. Since the responses of retrieval-based are generally more natural, fluent, diverse, and syntactically correct, retrieval-based method, which is our major focus in this work, is more mature for producing multi-turn dialogue systems both in academia and industry \cite{shum2018eliza,zhu-etal-2018-lingke,li2017alime}. Table \ref{tab:mutual} shows an example from Multi-Turn Dialogue Reasoning dataset (MuTual) \cite{cui2020mutual}. To choose the right answer, machine should understand and infer from the meaning of \textit{``table"} and the coreference, indicating the requirement of reasoning ability instead of simple matching.

\begin{table}
	\centering
	\resizebox{\linewidth}{!}
	{
		\begin{tabular}{l c l}
		    \toprule
		    Turn-ID & Speaker & Utterance (Context) \\
			 \midrule utterance 1 & \textit{F:} & \textit{Excuse me, sir. This is a non smoking area.} \\
			\midrule utterance 2 & \textit{M:} & \textit{Oh, sorry. \underline{I} will move to the smoking area.} \\
			\midrule utterance 3 & \textit{F:} & \textit{I’m afraid no \underline{table} in the smoking area is}\\ 
			& & \textit{available now}\\
			\midrule & & Response Candidates\\
			\midrule response 1 & \textit{M:} & \textit{Sorry. I won’t smoke in the \underline{hospital} again. \xmark} \\
			\midrule response 2 & \textit{M:} &  \textit{OK. I won't smoke. Could you please give me} \\
			& & \textit{a \underline{menu}? \checkmark} \\
			\midrule response 3 & \textit{M:} &  \textit{Could you please tell the \underline{customer over there}} \\
			& & \textit{not to \underline{smoke}? We can't stand the smell \xmark} \\
			\midrule response 4 & \textit{M:} & \textit{Sorry. I will smoke when I get off the \underline{bus}. \xmark}\\
			\bottomrule
		\end{tabular}
	}
		\caption{\label{tab:mutual} An example of response-selection for multi-turn dialogue in MuTual dataset.
	}
\end{table}

The core of previous mainstream work is to design a matching network, which calculates matching matrices for each utterance-response pair at different granularities. The matching matrices will be fused to get a feature vector, then the sequence of feature vectors will be further integrated by RNNs to get the final representation for scoring. However, these methods have obvious disadvantages:

$\bullet$ Interactions mainly involve each utterance and response, ignoring the global interactions between utterances.

$\bullet$ The relative positions between the response and different utterances are not taken into consideration, lacking the sequential information of context-response pairs.

Recently, pre-trained language models (PrLMs), such as BERT \cite{devlin2018bert}, RoBERTa \cite{liu2019roberta}, and ELECTRA \cite{clark2020electra}, have achieved impressive performance in a wide range of NLP tasks \cite{zhang2020mrc,zhang2019explicit,zhang2020SemBERT,zhang2019sgnet,li2020explicit,li2020global,li2019dependency}. In the multi-turn dialogue scope, there are also some efforts using PrLMs to yield promising performances \cite{whang2019domain,henderson2019training}. They are common in adopting a simple strategy by concatenating the response and context directly, and then adding classifier on the top layer of PrLMs \cite{zhou2020limit,zhang2019acl,zhang2021retro,xu2021topic}. However, such simple methods also have disadvantages:

$\bullet$ Simply embedding the token to high-dimensional space cannot faithfully model the additional information, such as positional or turn order information \cite{wang2019encoding}.

$\bullet$ The mechanism of self-attention runs through the whole dialogue, resulting in entangled information that originally belongs to different parts. Some work in other tasks made an effort to tackle this problem by explicitly separating different parts with self-attention and cross-attention, and achieve certain effects \cite{zhang2019dcmn+, zhu2020dual}.

From a broader view, the existing multi-turn dialogue methods have some common weaknesses. To our best knowledge, they rarely considers the following issues:

$\bullet$ \textbf{Transition}: There are speaker role transitions in multi-turn dialogue, which will lead to the ambiguity of coreference. Therefore, it is the essential difference between dialogue and passage. Although Speaker-Aware BERT \cite{gu2020speaker} has taken role transitions into consideration, adding the embedding of speaker-id into the inputs of PrLMs is not effective enough. 

$\bullet$ \textbf{Inherency}: Utterances have their own inherent meaning and contextualized meaning. Explicitly understanding local and global meaning respectively will reduce the negative effects of irrelevant contents in distant contexts. Some work on Neural Machine Translation has considered this and achieved promising results \cite{zheng2020toward}, but similar ideas have not appeared in multi-turn dialogue modeling.

In this work, we proposed a novel end-to-end Mask-based Decoupling-Fusing Network (MDFN) to tackle the above problems and thus fill the obvious gap of utterance-aware and speaker-aware representations. In detail, contextualized words representations of the dialogue texts are decoupled into four parts containing different information, which are then fused again after sufficient interactions. More specifically, the PrLM receives concatenated context and response, and outputs the contextualized representation of each word. For each word, we use a masking mechanism inside self-attention network, to limit the focus of each word only on the words in current utterance, other utterances, utterances of sender and utterances of receiver, respectively. To avoid ambiguity, we call the first two complementary parts as the utterance-aware channel, and the last two as the speaker-aware channel. We then fuse the information inside each channel via a gating mechanism to control the information reservation. The word-level information will be further aggregated to utterance level. For utterance-level representations, BiGRU is adopted to get dialogue-level representations. Inspired by \citet{tao2019multi}, we put the information fusion of two channels at the end to get the final dialogue representation for classification. 

Experimental results on four public benchmark datasets show that the proposed model outperforms the baseline models substantially on all the evaluation metrics, and achieves various new state-of-the-art results. Especially, compared to the strong ELECTRA baseline, experimental results verify that our model can enhance its performance by 1.6\% R@1 on the MuTual dataset.


\section{Background and Related Work}
\subsection{Matching Models}
\label{sec:relatedwork}
Matching models aim to match the response with contexts and calculate the matching score. They can be divided into two categories: single-turn and multi-turn matching models. Earlier research mainly considered the context utterances as one single utterance, using the dual encoder framework to encode the whole context and the response respectively. The encoder varies in LSTM \cite{hochreiter1997long}, CNNs \cite{lecun1998gradient}, Attentive-LSTM \cite{tan2015lstm}. 

\begin{figure*}[!htb]
\centering
\includegraphics[width=2\columnwidth]{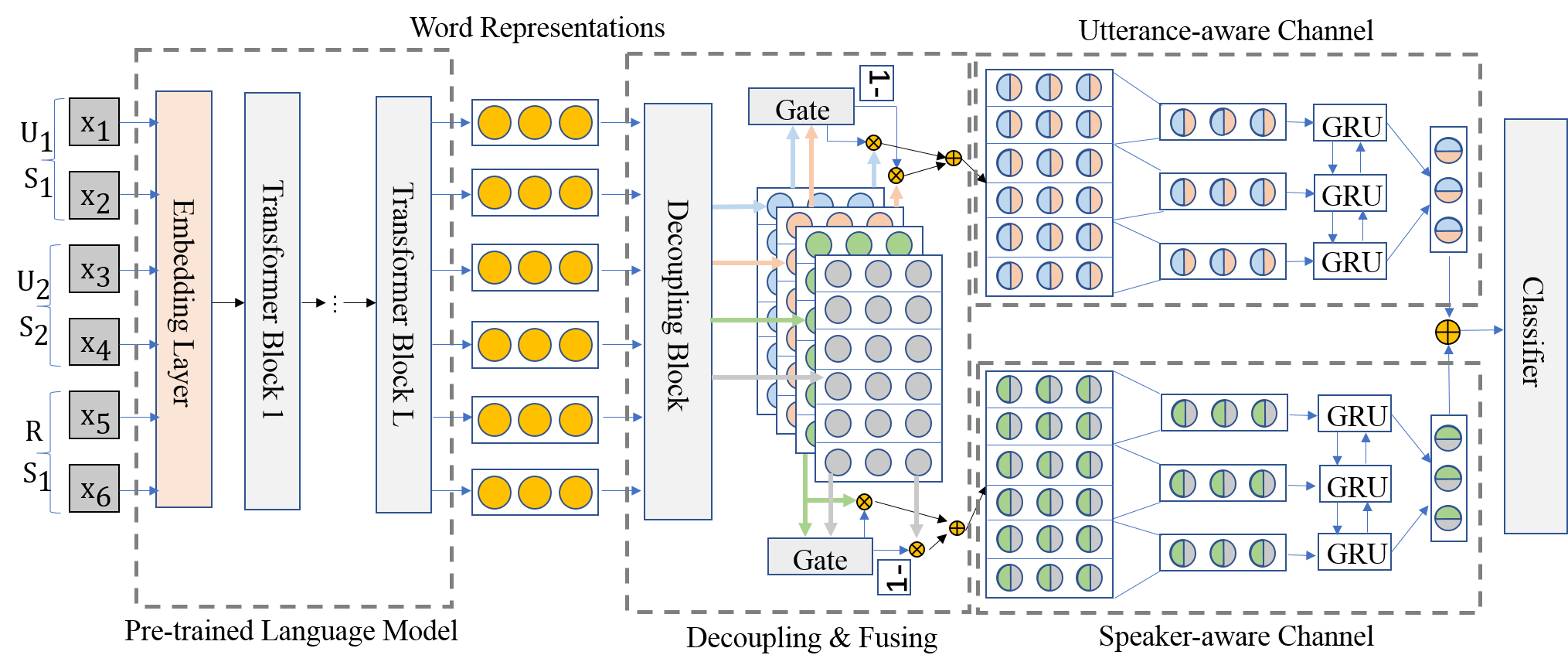} 
\caption{Overall framework of MDFN. Here for simplicity, inputs contain two utterances and one response spoken by two speakers with total six words. A more detailed figure and description on the decoupling block will be shown below.}
\label{fig:overall}
\end{figure*}
\begin{figure*}[!htb]
\centering
\includegraphics[width=1.93\columnwidth]{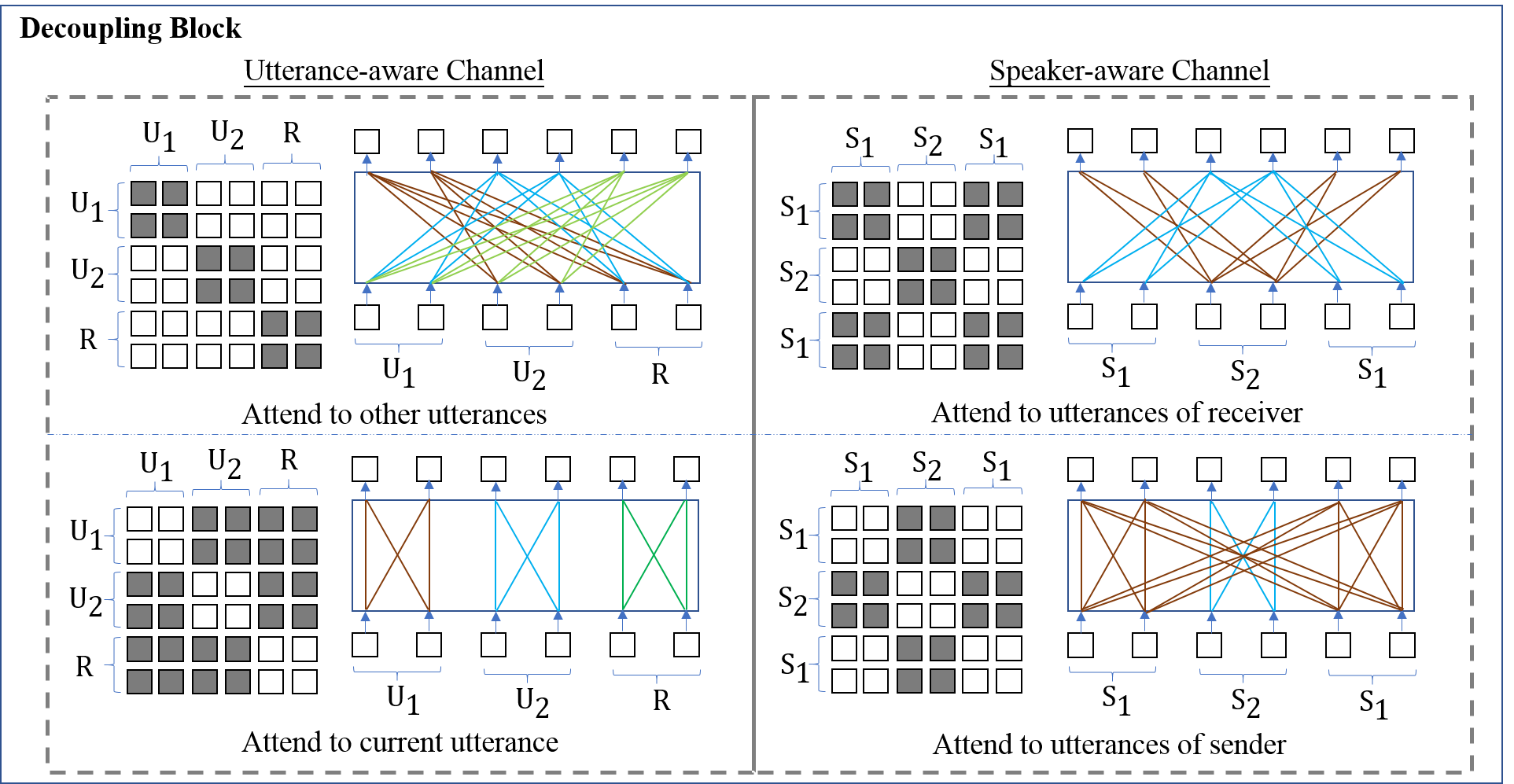} 
\caption{The decoupling block is composed of four independent self-attention blocks with the same inputs and different masks. The same word in different blocks attends to different scope of information}.
\label{fig:decoupling}
\end{figure*}

As for multi-turn matching, Sequential Matching Network (SMN) \cite{wu2016sequential} is the representative work, where a word-word and a sequence-sequence similarity matrix will be calculated, forming a matching image for each utterance-response pair which will be further integrated by CNN and RNN. Several related works like Deep Utterance Aggregation (DUA) \cite{zhang2018modeling}, Deep Attention Matching Network (DAM) \cite{zhou2018multi}, Interaction-over-Interaction (IoI) \cite{tao2019one} are extensions of SMN from different perspectives. DUA \cite{zhang2018modeling} points out that the last utterance should be considered explicitly and use self-attention to get a better segment matching matrix. DAM \cite{zhou2018multi} uses hierarchically stacked layers of self-attention to represent matching from different levels. IoI \cite{tao2019one} mainly deepens the layers of response-utterance interaction blocks and combines final states at different depths. 

\subsection{Pre-trained Language Models}
Most of pre-trained Language Models (PrLMs) are based on the encoder in Transformer, among which Bidirectional Encoder Representations from Transformers (BERT) \cite{devlin2018bert} is one of the most representative work. BERT uses multiple layers of stacked Transformer Encoder to obtain contextualized representations of the language at different levels. BERT has achieved unprecedented performances in many downstream tasks of NLP. Several subsequent variants have been proposed to further enhance the capacity of PrLMs, such as RoBERTa \cite{liu2019roberta}, ELECTRA \cite{clark2020electra}. Because of the good performance and fast inference speed of ELECTRA, we will adopt ELECTRA as our backbone model in this work.

\section{Mask-based Decoupling-Fusing Network}
Our proposed model, named Mask-based Decoupling-Fusing Network (MDFN), consists of six parts: Encoding, Decoupling, Fusing, Words Aggregating, Utterances Integrating and Scoring. In this part, we will formulate the problem and then introduce each sub-module in detail.
\subsection{Problem Formulation}
Suppose that we have a dataset $\mathcal{D} = \{(y_i, c_i, r_i)\}^N_{i=1}$, where $c_i = \{u_{i,1},...,u_{i,n_i}\}$ represents the dialogue context with $\{u_{i,k}\}^{n_i}_{k=1}$ as utterances. $r_i$ is a response candidate, and $y_i \in \{0,1\}$ denotes a label. The goal is to learn a discriminator $g(\cdot,\cdot)$ from $\mathcal{D}$, and at the inference phase, given any new context $c$ and response $r$, we use the discriminator to calculate $g(c,r)$ as their matching score. To select several best responses as return to human inputs, we rank matching scores or set a threshold among the list of candidates.
\subsection{Model Overview}

Figure \ref{fig:overall} shows the overall framework of MDFN. MDFN first encodes each word in the concatenated context and response to contextualized representation using the PrLMs. Then as shown in Figure \ref{fig:decoupling}, in the decoupling module, self-attention mechanism with different masks forces each word to only focus on the information in specific scopes, decoupling to utterance-aware and speaker-aware channel. Note that now the information flow inside different channels is independent. In each channel, complementary information will be fused by a gate. Then word representations belonging to the same utterance will be aggregated to represent the utterance. The sequence of utterance representations will be delivered to Bidirectional Gated Recurrent Unit (BiGRU) to produce the channel-aware dialogue-level representation. After that, the information from different channels is fused to get the final dialogue representation. The classifier will then output a score of the dialogue to reflect the matching degree between context and candidate response. 

MDFN is superior to existing methods in the following ways. First, compared to SA-BERT \cite{gu2020speaker}, the speaker-aware information is built upon semantically rich contextualized representations instead of a simple additional embedding term, making its influence on dialogue scoring preserved better. Second, compared to the work of designing a complicated matching network, our structure is simpler and can represent information at different levels in a more unified manner. Third, compared to pure PrLMs, additional layer of MDFN can decouple information to different channels, making a full use of the contextualized representations. Besides, since the number of utterances is far less than that of words, the top BiGRU layers can keep their memory capacity, alleviating the defects of Transformer on reflecting relative position information.  
\subsection{Encoding Context and Response}
\label{sec:PrLMs}
We first employ a pre-trained language model such as ELECTRA to obtain the initial word representations. The utterances and response are concatenated and then fed into the encoder. Note that we also insert ``[SEP]" between every two adjacent utterances. Let $U_i = [u_{i,0}, u_{i,1}, ..., u_{i,n_i}]$, $R = [r_0, r_1, ..., r_{n_r}]$, where $n_i$ and $n_r$ is respectively the length of $i$-th utterances and response, then the output is $ E = [e_1, e_2, ..., e_{n_0 + ... + n_k + n_r}]$ where $k$ is the maximum number of utterances, and $e_i \in \mathbf{R}^d$, $d$ is hidden size.
\subsection{Channel-aware Information Decoupling}
We first introduce the multi-head self-attention (MHSA) with mask, which can be formulated as:
\begin{equation*}
\begin{split}
    & \text{Attention}(Q,K,V,M) = \text{softmax}(\dfrac{QK^T}{\sqrt{d_k}} + M)V, \\ 
    & \text{head}_i = \text{Attention}(EW_i^Q,EW_i^K,EW_i^V,M),\\
    & \text{MHSA}(E,M) = \text{Concat}(\text{head}_1,...,\text{head}_h)W^O,
\end{split}
\end{equation*}
where $W_i^Q \in \mathbb{R}^{d_{model}\times d_q}$, $W_i^K \in \mathbb{R}^{d_{model} \times d_k}$, $W_i^V \in \mathbb{R}^{d_{model} \times d_v}$, $W_i^O \in \mathbb{R}^{hd_v \times d_{model}}$ are parameter matrices, $d_q$, $d_k$, $d_v$ denote the dimension of Query vectors, Key vectors and Value vectors, $h$ denotes the number of heads. $M$ denotes the mask.

We have four masks $\{M_k\}_{k=1}^4 \in \mathbb{R}^{l \times l}$ defined as:
\begin{equation*}
    \begin{split}
        M_1[i, j] &=\left\{\begin{array}{cc}
0, & \text { if } \mathbb{T}_{i} = \mathbb{T}_{j} \\
-\infty, & \text { otherwise }
\end{array}\right.\\
        M_2[i, j] &=\left\{\begin{array}{cc}
0, & \text { if } \mathbb{T}_{i} \neq \mathbb{T}_{j} \\
-\infty, & \text { otherwise }
\end{array}\right.\\
        M_3[i, j] &=\left\{\begin{array}{cc}
0, & \text { if } \mathbb{S}_{i} = \mathbb{S}_{j} \\
-\infty, & \text { otherwise }
\end{array}\right.\\
        M_4[i, j] &=\left\{\begin{array}{cc}
0, & \text { if } \mathbb{S}_{i} \neq \mathbb{S}_{j} \\
-\infty, & \text { otherwise }
\end{array}\right.\\
    \end{split}
\end{equation*}
where $i$, $j$ is the word position in the whole dialogue, $\mathbb{T}_i$ is the index of utterance the $i$-th word is located in, and $\mathbb{S}_i$ is the speaker the $i$-th word is spoken by. And $M_1$, $M_2$, $M_3$, $M_4$ only attends to the word in current utterance, other utterances, utterances of sender, and utterances of receiver. We call the first two utterance-aware channel, and the last two speaker-aware channel.

Then as shown in Figure \ref{fig:decoupling}, the decoupled channel-aware information $\{C_k\}_{k=1}^4 \in \mathbb{R}^{l \times d}$ are derived by multi-head self-attention with mask. 
\begin{equation}
\label{eq:CI}
    C_i = \text{MHSA}(E, M_i), i\in \{1,2,3,4\},
\end{equation}
where $E \in \mathbb{R}^{l \times d}$ is the output of PrLMs stated in Section \ref{sec:PrLMs}.
\subsection{Complementary Information Fusing}
To fuse the complementary information inside each channel, we use a gate to fuse them. Inspired by \cite{mou2015natural}, we use a gate to calculate the ratio of information preservation based on a matching heuristics considering the ``Siamese'' architecture as information from two parts, the element-wise product and difference as “similarity” or “closeness” measures. Let $E$ denote the original representations output from PrLMs. For utterance-aware channel, $\bar{E} = C_1$ and $\hat{E} = C_2$. For speaker-aware channel, $\bar{E} = C_3$ and $\hat{E} = C_4$. $C_i$ is defined in Equation (\ref{eq:CI}). The gate is formulated as:
\begin{equation}
    \label{eq:gate}
    \begin{split}
        &\Tilde{E_1} = \text{ReLU}(\text{FC}([E,\bar{E}, E-\bar{E}, E \odot \bar{E}])), \\
        &\Tilde{E_2} = \text{ReLU}(\text{FC}([E,\hat{E}, E-\hat{E}, E \odot \hat{E}])), \\
        &P = \text{Sigmoid}(\text{FC}(([\Tilde{E_1}, \Tilde{E_2}]))), \\
        &G(E, \bar{E}, \hat{E}) = P, \\
    \end{split}
\end{equation}
where Sigmoid, ReLU \cite{agarap2018deep} are activation functions, FC is fully-connected layer, and [$\cdot,\cdot$] means concatenation.

Using two parametric-independent gates, the channel-aware information can be fused, which is defined as:
\begin{equation*}
    \begin{split}
        & P_1 = G_1(E,C_1,C_2), \\
        & P_2 = G_2(E,C_3,C_4), \\
        & C_u = P_1 \odot C_1 + (1-P_1)  \odot C_2, \\
        & C_s = P_2 \odot C_3 + (1-P_2)  \odot C_4,
    \end{split}
\end{equation*}
where the calculation of $\{G_i\}_{i=1}^2$ is defined in Equation \ref{eq:gate}, $C_u$ and $C_s$ is the fused utterance-aware and speaker-aware word representations, respectively.
\subsection{Utterance Representations}
For each channel, word representations will be aggregated by simple max-pooling over words in the same utterance to get the utterance representations. Let $L_u$ and $L_s$ be the output in this part. Then they are defined as:
\begin{equation*}
\begin{split}
    & L_u[i,:] = \mathop{\text{MaxPooling}}\limits_{\mathbb{T}_j=i}({C_u[j,:]}) \in \mathbb{R}^d, \\
    & L_s[i,:] = \mathop{\text{MaxPooling}}\limits_{\mathbb{T}_j=i}({C_s[j,:]}) \in \mathbb{R}^d.
\end{split}
\end{equation*}
\subsection{Dialogue Representation}
To get the channel-aware dialogue representation, the sequence of utterance representations will be delivered to BiGRU. Suppose that the hidden states of the BiGRU are $(\boldsymbol{h}_1, \dots, \boldsymbol{h}_k)$, then $\forall j$, $1\leq j \leq k$, $\boldsymbol{h}_j \in \mathbb{R}^{2d}$ is given by
\begin{equation*}
    \begin{split}
        &\overleftarrow{\boldsymbol{h}}_j = \overleftarrow{\text{GRU}}(\overleftarrow{\boldsymbol{h}}_{j-1}, \overleftarrow{\boldsymbol{L}}[j]) ,\\
        &\overrightarrow{\boldsymbol{h}}_j = \overrightarrow{\text{GRU}}(\overrightarrow{\boldsymbol{h}}_{j-1}, \overrightarrow{\boldsymbol{L}}[j]) ,\\
        & \boldsymbol{h}_j = [\overleftarrow{\boldsymbol{h}}_j; \overrightarrow{\boldsymbol{h}}_j].
    \end{split}
\end{equation*}
For each channel, we take the hidden state of BiGRU at the last step as channel-aware dialogue representation. Let the two vectors be $\boldsymbol{v_1}$ and $\boldsymbol{v_2}$. Then two channels can be easily fused to get the final dialogue representation.
\begin{equation*}
    \boldsymbol{v} = \text{Tanh}(W[\boldsymbol{v_1};\boldsymbol{v_2}] + b),
\end{equation*}
where $W \in \mathbb{R}^{d \times 4d}$, $b \in \mathbb{R}^{d}$ are trainable parameters. Tanh is the activation function.
\subsection{Scoring and Learning}
The dialogue vector will be fed into a classifier with a fully connected and softmax layer. We learn model $g(\cdot, \cdot)$ by minimizing cross entropy loss with dataset $\mathcal{D}$. Let $\Theta$ denote the parameters of MDFN, for binary classification like ECD, Douban and Ubuntu, the objective function $\mathcal{L(D}, \Theta)$ can be formulated as:
\begin{equation*}
    -\sum_{i=1}^N [y_ilog(g(c_i,r_i)) + (1-y_i)log(1-g(c_i,r_i))].
\end{equation*}
For multiple choice task like MuTual, the loss function is:
\begin{equation*}
    -\sum_{i=1}^N\sum_{k=1}^C y_{i,k}log(g(c_i,r_{i,k})).
\end{equation*}

\begin{table*}
{
    \centering
    \small
     \setlength{\tabcolsep}{4pt}
    {
        \renewcommand\arraystretch{1.1}
        \begin{tabular}{l|cccc|cccccc|ccc}
            \hline \textbf{Model} & \multicolumn{4}{c|}{\textbf{Ubuntu Corpus}} & \multicolumn{6}{c|}{\textbf{Douban Conversation Corpus}} & \multicolumn{3}{c}{\textbf{E-commerce Dialogue Corpus}} \\
            \cline{2-14} & $\textbf{R}_{2}$@1 & $\textbf{R}_{10}$@1 & $\textbf{R}_{10}$@2 & $\textbf{R}_{10}$@5 & \textbf{MAP} & \textbf{MRR} & \textbf{P}@1 & $\textbf{R}_{10}$@1 & $\textbf{R}_{10}$@2 & $\textbf{R}_{10}$@5 & $\textbf{R}_{10}$@1 & $\textbf{R}_{10}$@2 & $\textbf{R}_{10}$@5 \\
            
            \hline SMN & 0.926 & 0.726 & 0.847 & 0.961 & 0.529 & 0.569 & 0.397 & 0.233 & 0.396 & 0.724 & 0.453 & 0.654 & 0.886 \\
             DUA & - & 0.752 & 0.868 & 0.962 & 0.551 & 0.599 & 0.421 & 0.243 & 0.421 & 0.780 & 0.501 & 0.700 & 0.921 \\
             DAM & 0.938 & 0.767 & 0.874 & 0.969 & 0.550 & 0.601 & 0.427 & 0.254 & 0.410 & 0.757 & - & - & - \\
             IoI & 0.947 & 0.796 & 0.894 & 0.974 & 0.573 & 0.621 & 0.444 & 0.269 & 0.451 & 0.786 & - & - & - \\
             MSN & - & 0.800 & 0.899 & 0.978 & 0.587 & 0.632 & 0.470 & 0.295 & 0.452 & 0.788 & 0.606 & 0.770 & 0.937 \\
             MRFN & 0.945 & 0.786 & 0.886 & 0.976 & 0.571 & 0.617 & 0.448 & 0.276 & 0.435 & 0.783 & - & - & - \\
             \hline BERT & 0.950 & 0.808 & 0.897 & 0.975 & 0.591 & 0.633 & 0.454 & 0.280 & 0.470 & 0.828 & 0.610 & 0.814 & 0.973 \\
             SA-BERT & 0.965 & 0.855 & 0.928 & 0.983 & 0.619 & 0.659 & 0.496 & 0.313 & 0.481 & 0.847 & \textbf{0.704} & \textbf{0.879} & \textbf{0.985} \\
             \hdashline
             ELECTRA & 0.960 & 0.845 & 0.919 & 0.979 & 0.599 & 0.643 & 0.471 & 0.287 & 0.474 & 0.831 & 0.607 & 0.813 & 0.960 \\
            \textbf{MDFN} & \textbf{0.967} & \textbf{0.866} & \textbf{0.932} & \textbf{0.984} &  \textbf{0.624} & \textbf{0.663} & \textbf{0.498} & \textbf{0.325} & \textbf{0.511} & \textbf{0.855} & 0.639 &  0.829 & 0.971 \\
             \hline
        \end{tabular}
    }
    \caption{\label{tab:ubuntu} Results on three benchmark datasets. The upper part includes methods of multi-turn matching network, and the lower part includes methods based on PrLMs. ELECTRA is our implemented baseline.
	}
}
\end{table*}

\begin{table}[!btb]
{
    \centering\small
    \setlength{\tabcolsep}{12pt}
    {
        \begin{tabular}{lccc}
            \hline \textbf{Model} & $\textbf{R}_{4}$@1 & $\textbf{R}_{4}$@2 & MRR \\
            \hline \emph{On Leaderboard} &  \\
             RoBERTa & 0.713 & 0.892 & 0.836  \\
             DRRC-1 & 0.771 & 0.914 & 0.869 \\
             RoBERTa+ & 0.825 & 0.953 & 0.904 \\
             RoBERTa+OCN & 0.867 & 0.958 & 0.926 \\
             UMN & 0.870 & 0.973 & 0.930 \\
             RoBERTa++ & 0.903 & 0.976 & 0.947 \\
             UGCN & 0.915 & 0.983 & 0.954 \\
            \hline \emph{In Literature} &  \\
             
             SMN & 0.264 & 0.524 & 0.578 \\
             DAM & 0.261 & 0.520 & 0.645 \\
             BERT & 0.514 & 0.787 & 0.715 \\
             \hline
             ELECTRA & 0.900 & 0.979 & 0.946 \\
             \textbf{MDFN} & \textbf{0.916} & \textbf{0.984} & \textbf{0.956} \\
             \hline
        \end{tabular}
    }
    \caption{\label{tab:mutual_result} Results on MuTual dataset.
	}
}
\end{table}

\section{Experiments}
\subsection{Datasets}
We tested our model\footnote{The implementation of our MDFN model is available at \url{https://github.com/comprehensiveMap/MDFN}} on two English datasets: Ubuntu Dialogue Corpus (Ubuntu) and Multi-Turn Dialogue Reasoning (MuTual), and two Chinese datasets: Douban Conversation Corpus (Douban) and E-commerce Dialogue Corpus (ECD).
\subsubsection{Ubuntu Dialogue Corpus} Ubuntu \cite{lowe2015ubuntu} consists of English multi-turn conversations about technical support collected from chat logs of the Ubuntu forum. The dataset contains 1 million context-response pairs, 0.5 million for validation and 0.5 million for testing. In training set, each context has one positive response generated by human and one negative response sampled randomly. In validation and test sets, for each context, there are 9 negative responses and 1 positive response. 
\subsubsection{Douban Conversation Corpus} Douban \cite{wu2016sequential} is different from Ubuntu in the following ways. First, it is open domain where dialogues are extracted from Douban Group. Second, Response candidates on the test set are collected by using the last turn as query to retrieve 10 response candidates and labeled by humans. Third, there could be more than one correct responses for a context.
\subsubsection{E-commerce Dialogue Corpus} ECD \cite{zhang2018modeling} dataset is extracted from conversations between customer and service staff on Taobao. It contains over 5 types of conversations based on over 20 commodities. There are also 1 million context-response pairs in training set, 0.5 million in validation set and 0.5 million in test set.
\subsubsection{Multi-Turn Dialogue Reasoning} MuTual \cite{cui2020mutual} consists of 8860 manually annotated dialogues based on Chinese student English listening comprehension exams. For each context, there is one positive response and three negative responses. The difference compared to the above three datasets is that only MuTual is reasoning-based. There are more than 6 types of reasoning abilities reflected in MuTual.
\subsection{Setup}
For the sake of computational efficiency, the maximum number of utterances is specialized as 20. The concatenated context, response, ``[CLS]" and ``[SEP]" in one sample is truncated according to the ``longest first" rule or padded to a certain length, which is 256 for MuTual and 384 for the other three datasets. Our model is implemented using Pytorch and based on the Transformer Library. We use ELECTRA \cite{clark2020electra} as our underlying model. AdamW \cite{loshchilov2017decoupled} is used as our optimizer. The batch size is 24 for MuTual, and 64 for others. The initial learning rate is $4\times 10^{-6}$ for MuTual and $3\times 10^{-6}$ for others. We run 3 epochs for MuTual and 2 epochs for others and select the model that achieves the best result in validation. For the English tasks, we use the pre-trained weights \textit{electra-large-discriminator} for fine-tuning; for the Chinese tasks, the weights are from \textit{hfl/chinese-electra-large-discriminator}.\footnote{Those weights are available in the Transformers repo: \url{https://github.com/huggingface/transformers/}.}

\subsection{Baseline Models}
The following models are our baselines for comparison:

$\bullet$ \textbf{Multi-turn matching models}: Sequential Matching Network (SMN) \cite{wu2016sequential}, Deep Attention Matching Network (DAM) \cite{zhou2018multi}, Deep Utterance Aggregation (DUA) \cite{zhang2018modeling}, Interaction-over-Interaction (IoI) \cite{tao2019one} have been stated in Section \ref{sec:relatedwork}. Besides, Multi-Representation Fusion Network (MRFN) \cite{tao2019multi} matches context and response with multiple types of representations. Multi-hop Selector Network (MSN) \cite{yuan2019multi} utilizes a multi-hop selector to filter necessary utterances and match among them.

$\bullet$ \textbf{PrLMs-based models}: BERT \cite{devlin2018bert}, RoBERTa \cite{liu2019roberta}, ELECTRA \cite{clark2020electra} have been stated in Section \ref{sec:relatedwork}. Besides, Option Comparison Network (OCN) \cite{ran2019option} is involved, which compares the options before matching response and contexts.\footnote{ On the leaderboard of MuTual, since some work is not publicly available, we will not introduce here.}

\begin{table}
		\centering
        \small
        \setlength{\tabcolsep}{8pt}
    {
		{
			\begin{tabular}{l c c c}
				\hline
				\textbf{Model} &\textbf{R@1} & \textbf{R@2} & \textbf{MRR} \\
				\hline
				ELECTRA & 0.903   & 0.980 & 0.948 \\
				\quad + UA-Mask & 0.909 & 0.973 & 0.949 \\
				\quad + SA-Mask & 0.913 & 0.977 & 0.953 \\ 
				\hdashline
				\quad + MDFN & \textbf{0.923} & 0.979 & \textbf{0.958} \\
				\quad\quad - BiGRU + Max-Pool & 0.909 & \textbf{0.982} & 0.951 \\
				\quad\quad - BiGRU + Mean-Pool  & 0.911  & 0.980 & 0.952 \\    
				\hline
			\end{tabular}
		}
    }
		\caption{\label{tab:ablation} Ablation study on MuTual dev sets. ``UA" means utterance-aware, and ``SA" means speaker-aware.}
	\end{table}
	
\subsection{Evaluation Metrics}
Following the previous work \cite{lowe2015ubuntu, wu2016sequential}, we calculate the proportion of true positive response among the top-$k$ selected responses from the list of $n$ available candidates for one context, denoted as $\textbf{R}_n$@$k$. Besides, additional conventional metrics of information retrieval are employed on Douban: Mean Average Precision (MAP) \cite{baeza1999modern}, Mean Reciprocal Rank (MRR) \cite{voorhees1999trec}, and precision at position 1 (P@1). 

\subsection{Results}
Tables \ref{tab:ubuntu}-\ref{tab:mutual_result} show the results on four datasets. Our model outperforms other models in most metrics. Generally, the previous models based on multi-turn matching network perform worse than simple PrLMs-based ones, illustrating the power of contextualized representations in context-sensitive dialogue modeling. PrLM can perform even better when equipped with MDFN, verifying the effectiveness of our model, where utterance-aware and speaker-aware information can be better exploited. In particular, our model surpasses the previous SOTA SA-BERT model on most datasets, which is augmented by extra domain adaptation strategies to conduct language model pre-training on in-domain dialogue corpus before fine-tuning on tasks. In addition, MDFN also ranks the best on the MuTual leaderboard.\footnote{https://nealcly.github.io/MuTual-leaderboard/}

\section{Analysis}

\begin{table}
		\centering
        \small
        \setlength{\tabcolsep}{7pt}
    {
		{
			\begin{tabular}{c c c c c c}
				\hline
				\textbf{\#Decoupling} & 1 & 2 & 3 & 4 & 5 \\
				\hline
				\textbf{R@1} & \textbf{0.923} & 0.909 & 0.910 & 0.913 & 0.911\\
				\textbf{R@2} & \textbf{0.979} & 0.974 & 0.974 & 0.976 & 0.978\\
				\textbf{MRR} & \textbf{0.958} & 0.950 & 0.950 & 0.952 & 0.951\\ 
				\hline
			\end{tabular}
		}
    }
		\caption{\label{tab:Decouple} Influence of the number of Decoupling Blocks.}
	\end{table}

\begin{table}
		\centering\small
        \setlength{\tabcolsep}{6pt}
    {
		{
			\begin{tabular}{l c c c c c}
				\hline
				\textbf{\#BiGRU Layer} & 1 & 2 & 3 & 4 & 5 \\
				\hline
				\textbf{R@1} & \textbf{0.923} & 0.906 & 0.912 & 0.901 & 0.903 \\
				\textbf{R@2} & 0.979 & 0.974 & \textbf{0.980} & 0.979 & 0.973 \\
				\textbf{MRR} & \textbf{0.958} & 0.949 & 0.952 & 0.946 & 0.946 \\ 
				\hline
			\end{tabular}
		}
        }
		\caption{\label{tab:RNN} Influence of the number of BiGRU layers.}
	\end{table}

\begin{table}
		\centering\small
			\setlength{\tabcolsep}{13pt}
		{
			\begin{tabular}{l c c c}
				\hline
				\textbf{Fusing Method} & \textbf{R@1} & \textbf{R@2} & \textbf{MRR} \\
				\hline
				\textbf{MDFN} & \textbf{0.923} & 0.979 & \textbf{0.958} \\
				\quad \textbf{-Gate} & 0.914 & 0.975 & 0.952 \\
				\quad \textbf{-Original Info} & 0.918 & 0.980 & 0.955 \\ 
				\quad \quad \textbf{-Gate} & 0.910 & \textbf{0.981} & 0.951 \\ 
				\hline
			\end{tabular}
		}
		\caption{\label{tab:Gate} Influence of fusing methods.}
	\end{table}
	
\subsection{Ablation Study}
Since our model decouples information to utterance-aware and speaker-aware channels, and uses BiGRU to learn a channel-aware dialogue representation from the sequence of utterance representations. We wonder the effect of two channels and whether BiGRU can be replaced to simple pooling. We perform an ablation study on MuTual dev sets as shown in Table \ref{tab:ablation}. Results show that each part is essentially necessary. The most important part is speaker-aware information, since speaker role transition is an essential feature in multi-party dialogues. Then it comes to BiGRU, revealing the strength of BiGRU when modeling a short sequence.

\subsection{Number of Decoupling Layers}
Inspired by the stacked manner of Transformer \cite{vaswani2017attention}, we have also done a study on the effects of stacked decoupling layers. The results are shown on Table \ref{tab:Decouple}, where we can see only one layer is enough and better than deeper. This can be explained that a deeper decoupling module is harder to learn and only one interation is enough.

\subsection{Number of BiGRU Layers}
To see whether a deeper BiGRU will be beneficial to the modeling of a dialogue representation from the sequence of utterance representations, we have conducted experiments on the number of BiGRU layers. As shown in Table \ref{tab:RNN}, a deeper BiGRU causes a big drop on \textbf{R}@1 metric, showing that for a short sequence which is already contextualized, shallow BiGRU is enough.

\subsection{Effects of Fusing Methods}
To fuse the complementary information inside one channel, we can use gate to decide their preservation ratio. Another intuitive way is just using a simple fully-connected layer to fuse them directly instead of calculating the ratio. And our gate calculation considers the original word representations as shown in Equation (\ref{eq:gate}). Here, we explore the effects of whether using gate or whether considering original information. The results are shown in Table \ref{tab:Gate}. We can see the gate mechanism is necessary because the two kinds of information are inherently complementary. Besides, since decoupling information can be seen as a transformation from original information, the original one can be a good reference while gate ratio calculating.

\subsection{Effects of Aggregating Methods}
To aggregate the word representations in one utterance, we can use simple global pooling. And another widely used method is Convolution Neural Network (CNN). Deeper CNN can capture a wide scope of receptive field, making it successful in Computer Vision \cite{simonyan2014very} and Text Classification \cite{kim2014convolutional}. We have also compared different aggregating methods on MuTual dev set as shown in Table \ref{tab:aggregating}, we can see max-pooling is better than mean-pooling, since it can preserve some activated signals, removing the disturbance of less important signals. However, the two CNN-based methods perform worse than max-pooling, especially that with multiple filter sizes. This can be explained that shared filters between different sentences is not flexible enough and cannot generalize well.
\begin{table}
		\centering\small
		\setlength{\tabcolsep}{12pt}
		{
			\begin{tabular}{l c c c}
				\hline
				\textbf{Aggregating Method} & \textbf{R@1} & \textbf{R@2} & \textbf{MRR} \\
				\hline
				\textbf{Max-Pooling} & \textbf{0.923} & \textbf{0.979} & \textbf{0.958} \\
				\textbf{Mean-Pooling} & 0.911 & 0.975 & 0.951 \\
				\textbf{CNN} & 0.916 & 0.974 & 0.953 \\ 
				\textbf{CNN-Multi} & 0.902 & 0.974 & 0.946 \\ 
				\hline
			\end{tabular}
		}
		\caption{\label{tab:aggregating} Influence of the Aggregating Methods. CNN uses filter of size 3, and CNN-Multi combines filters of size 2, 3, 4, which is similar to \cite{kim2014convolutional}.}
	\end{table}	

\subsection{Effects of Underlying Pre-trained Models}
To test the generality of the benefits our MDFN to other PrLMs, we alter the underlying PrLMs to other variants in different sizes or types. As shown in Table \ref{tab:PrLMs}, 
we see that our MDFN is generally effective to the widely-used PrLMs.

\begin{table}
		\centering\small
\setlength{\tabcolsep}{1.3pt}
    {
		{
			\begin{tabular}{l c c c c c c}
				\hline
				\textbf{PrLMs} & \multicolumn{2}{c}{\textbf{R@1}} & \multicolumn{2}{c}{\textbf{R@2}} & \multicolumn{2}{c}{\textbf{MRR}} \\
				 & Single & \textit{+MDFN} & Single & \textit{+MDFN} & Single & \textit{+MDFN} \\
				\hline
				\textbf{$\text{BERT}_\text{base}$} & 0.653 & \textbf{0.684} & 0.860 & \textbf{0.871} & 0.800 & \textbf{0.818} \\
				\textbf{$\text{RoBERTa}_\text{base}$} & 0.709 & \textbf{0.731} & 0.886 & \textbf{0.898} & 0.833 & \textbf{0.846} \\
				\textbf{$\text{ELECTRA}_\text{base}$} & 0.762 & \textbf{0.813} & 0.916 & \textbf{0.928} & 0.865 & \textbf{0.893} \\
				\textbf{$\text{BERT}_\text{large}$} & 0.691 & \textbf{0.726} & 0.879 & \textbf{0.901} & 0.822 & \textbf{0.844} \\
				\textbf{$\text{RoBERTa}_\text{large}$} & 0.834 & \textbf{0.845} & 0.952 & \textbf{0.953} & 0.908 & \textbf{0.914} \\
				\textbf{$\text{ELECTRA}_\text{large}$} & 0.906 & \textbf{0.923} & 0.977 & \textbf{0.979} & 0.949 & \textbf{0.958} \\
				\hline
			\end{tabular}
		}}
		\caption{\label{tab:PrLMs} Performances of single PrLM and with MDFN.}
	\end{table}

\section{Conclusion}
In this paper, we propose a novel and simple Maked-based Decoupling-Fusing Network (MDFN), which decouples the utterance-aware and speaker-aware information, tackling the problem of role transition and noisy distant texts. Experiments on four retrieval-based multi-turn dialogue datasets show the superiority over existing methods. The ablation study of different sub-modules explains the effectiveness and relative importance of them. Our work reveals a way to make better use of the semantically rich contextualized representations from pre-trained language models and gives insights on how to combine the traditional RNN models with powerful transformer-based models. In the future, we will research further on other NLP tasks using a similar framework of MDFN to test its universality.

\bibliography{newrefs}
\end{document}